\pdfoutput=1

\documentclass[11pt]{article}

\usepackage{acl}%

\usepackage{times}
\usepackage{latexsym}

\usepackage[T1]{fontenc}

\usepackage[utf8]{inputenc}

\usepackage{microtype}

\usepackage{algorithm}
\usepackage{algorithmic}
\usepackage{amsmath,amssymb}
\usepackage{bm}
\usepackage{booktabs}
\usepackage{bussproofs}
\usepackage{color}
\usepackage{comment}
\usepackage{enumerate}
\usepackage{mathtools}
\usepackage{multirow}
\usepackage{natbib}
\usepackage{stmaryrd}
\usepackage{caption} %
\usepackage{subcaption} %
\usepackage{url}
\usepackage{hyperref}

\newcommand{\Acc}[2]{\textsf{Acc}(#1, #2)}

\newcommand{\back}{\backslash}

\newcommand{\book}[1]{\textsc{book}(#1)}
\newcommand{\bryce}[1]{\textsc{bryce}(#1)}

\newcommand{\child}[1]{\textsc{child}(#1)}
\newcommand{\dcl}{\mathrm{[dcl]}}

\newcommand{\have}[1]{\textsc{have}(#1)}

\newcommand{\Subj}[2]{\textsf{Subj}(#1, #2)}

\newcommand{\True}[1]{\textsf{True}({#1})}

\title{Logical Inference for Counting on Semi-structured Tables}

\author{Tomoya Kurosawa \and Hitomi Yanaka \\
        The University of Tokyo\\
        \texttt{\{kurosawa-tomoya, hyanaka\}@is.s.u-tokyo.ac.jp}}

\begin{document}
\maketitle
\begin{abstract}
Recently, the Natural Language Inference (NLI) task has been studied for semi-structured tables that do not have a strict format.
Although neural approaches have achieved high performance in various types of NLI, including NLI between semi-structured tables and texts, they still have difficulty in performing a numerical type of inference, such as counting.
To handle a numerical type of inference, we propose a logical inference system for reasoning between semi-structured tables and texts.
We use logical representations as meaning representations for tables and texts and use model checking to handle a numerical type of inference between texts and tables.
To evaluate the extent to which our system can perform inference with numerical comparatives, we make an evaluation protocol that focuses on numerical understanding between semi-structured tables and texts in English.
We show that our system can more robustly perform inference between tables and texts that requires numerical understanding compared with current neural approaches.
\end{abstract}

\section{Introduction}

Natural Language Inference (NLI) \citep{dagan-etal-2005-rte} is one of the most fundamental tasks to determine whether a premise entails a hypothesis.
Recently, researchers have developed benchmarks not only for texts but for other kinds of resources as well, a table being one example.
Previous studies have targeted database-style structured tables \citep{pasupat-liang-2015-compositional,wiseman-etal-2017-challenges,krishnamurthy-etal-2017-neural} 
and semi-structured tables, such as the infoboxes in Wikipedia \citep{lebret-etal-2016-neural,gupta-etal-2020-infotabs}.
Our focus here is on the NLI task on semi-structured tables, where we handle a semi-structured table as a premise and a sentence as a hypothesis.

In Figure \ref{figure:joe-biden}, for example, we consider the semi-structured table as a given premise and take \textit{Joe Biden was born in November} as Hypothesis 1.
We can conclude that Hypothesis 1 is entailed by the table.
A semi-structured table has only two columns and describes a single object, which is indicated in the title.
We call elements of the first column, such as
\textbf{Political Party}, \textit{keys}, each of which has an associated \textit{value} in the second column such as \textit{Democratic} (\textit{1969--present}).
Pairs of keys and values are called \textit{rows}.
It is relatively difficult to understand the information contained in infobox tables because (i) values are not limited to words or phrases, and sometimes whole sentences, and (ii) a row can contain more than one type of information, such as the birthday and birthplace in the \textbf{Born} row.

\begin{figure}[t]
    \small
    \begin{center}
        \begin{tabular}{ll} \toprule
    \multicolumn{2}{c}{\textbf{Joe Biden}} \\ \midrule
    \textbf{Born} & Joseph Robinette Biden Jr. \\
    & November 20, 1942 (age 79) \\
    & Scranton, Pennsylvania, U.S. \\
    \textbf{Political party} & Democratic (1969--present) \\
    \textbf{Spouse(s)} & Neilia Hunter (m. 1966; died 1972) \\
    & Jill Jacobs (m. 1977) \\\bottomrule
\end{tabular}
    \end{center}
    \vspace{.5\baselineskip}

    \noindent
    Hypothesis 1: Joe Biden was born in November.\\
    Hypothesis 2: Joe Biden has had more than two wives.\\
    \caption{A semi-structured table describing Joe Biden\footnotemark and two hypothesis sentences.
    This table entails Hypothesis 1 and contradicts Hypothesis 2.
    }
    \label{figure:joe-biden}
\end{figure}

\footnotetext{The table was retrieved from \url{https://en.wikipedia.org/wiki/Joe_Biden} on February 25, 2022.
Some rows have been removed to save space.}

In recent years, modern neural network (NN) approaches have achieved high performance in many Natural Language Understanding benchmarks,
such as BERT \citep{devlin-etal-2019-bert}. 
NN-based approaches \cite{neeraja-etal-2021-incorporating} have also achieved high accuracy on the NLI task between semi-structured tables and texts,
but previous studies have questioned whether NN-based models truly understand the various linguistic phenomena \citep{jia-liang-2017-adversarial,naik-etal-2018-stress,rozen-etal-2019-diversify,ravichander-etal-2019-equate,Richardson_Hu_Moss_Sabharwal_2020}.
These studies have shown that NN-based approaches have failed to achieve a high performance in numerical reasoning. %

In this paper, we focus on a numerical type of inference on semi-structured tables, which requires understanding the number of items in a table as well as numerical comparisons.
Numerical comparatives are among the more challenging linguistic phenomena that involve generalized quantifiers. 
For example, the phrase \textit{more than} in Hypothesis 2 in Figure \ref{figure:joe-biden} is a numerical comparative and compares two and the number of wives. 
For dealing with numerical comparatives, \citet{haruta-etal-2020-combining,haruta-etal-2020-logical} achieved high performance by developing a logical inference system based on formal semantics.
However, \citet{haruta-etal-2020-combining,haruta-etal-2020-logical} concentrated on the inference between texts only, and inference systems that reliably perform inference between tables and texts involving numerical comparatives have not yet been developed.

Thus, we aim to develop a logical inference system between semi-structured tables and texts, especially for numerical reasoning.
While previous work~\citep{pasupat-liang-2015-compositional,wiseman-etal-2017-challenges,krishnamurthy-etal-2017-neural} has provided semantic parsers of constructing query languages such as SQLs for question answering on database-style tables,
we present logical representations for semi-structured tables to enable a numerical type of inference on semi-structured tables.
Furthermore, the existing NLI dataset for semi-structured tables \citep{gupta-etal-2020-infotabs} does not contain sufficient test cases for understanding numerical comparatives.
Thus, there is a need for an evaluation protocol that investigates the numerical reasoning skills of NLI systems for semi-structured tables.

Given this background, our main contributions in this paper are the following:
\begin{enumerate}[1.]
    \item We propose a logical inference system for handling numerical comparatives that is based on formal semantics for NLI between semi-structured tables and texts.
    \item We provide an evaluation protocol and dataset that focus on numerical comparatives between semi-structured tables and texts.
    \item We demonstrate the increased performance of our inference system compared with previous NN models on the NLI dataset, focusing on numerical comparatives between semi-structured tables and texts.
\end{enumerate}
Our system and dataset will be publicly available at \url{https://github.com/ynklab/sst_count}.

\section{Related Work and Background}
\label{chapter:related-work}
This section explains the related work of logic-based NLI approaches and the background of model checking, which is used for inference between semi-structured tables and sentences in our proposed system.

\subsection{Logic-based Approach}
Based on the analysis of formal semantics, logic-based NLI approaches can handle a greater variety of linguistic phenomena than NN-based approaches can.
Some logic-based NLI approaches using syntactic and semantic parsers based on formal semantics have been proposed \citep{bos-2008-wide,abzianidze-2015-tableau,mineshima-etal-2015-higher,hu-etal-2020-monalog,haruta-etal-2020-combining,haruta-etal-2020-logical}.
These logic-based approaches can derive semantic representations of sentences involving linguistically challenging phenomena, such as generalized quantifiers and comparatives, based on Combinatory Categorial Grammar (CCG) \citep{steedman-2000-ccg} syntactic analysis.
CCG is often used in these approaches because it has a tiny number of combinatory rules, which is suitable for semantic composition from syntactic structures.
In addition, robust CCG parsers are readily available \citep{clark-curran-2007-wide,yoshikawa-etal-2017-ccg}.

Regarding logic-based approaches for inference other than inference between texts, \citet{suzuki-etal-2019-multimodal} proposed a logical inference system for inference between images and texts.
Their system converts images to first-order logic (FOL) structures by using image datasets where structured representations of the images are annotated.
They then get FOL formulas $P$ for images from these structures along with the associated image captions.
Hypothesis sentences are translated into FOL formulas $H$ through the use of a semantic parser \cite{martinez-gomez-etal-2016-ccg2lambda}.
For inference, they used automated theorem proving and sought to prove $P\vdash H$.
Our proposed inference system between semi-structured tables and texts is inspired by \citet{suzuki-etal-2019-multimodal}.
While the previous system uses automated theorem proving for inference between images and texts, our system uses model checking to judge whether a given text is true under a given table, and it is expected to be a faster method.

\begin{figure}[t]
    \centering\small
    \input{figures/model_sample.tex}

    \vspace{-.5\baselineskip}
    \begin{tabular}{ll} \toprule
Logical formula & Output \\ \midrule
$\exists x. \exists y. (\textsc{boy}(x) \land \textsc{like}(x, y))$ & True \\
$\exists x. \exists y. (\textsc{girl}(x) \land \textsc{girl}(y) \land \textsc{like}(x, y))$ & False \\
$\exists x. \exists y. (\textsc{cat}(y) \land \textsc{like}(x, y))$ & Undefined \\ \bottomrule
\end{tabular}

    \caption{Outputs of model checking based on an example model and three formulas.}
    \label{figure:model-check-sample}
\end{figure}

\begin{figure*}[t]
    \centering
    \includegraphics[width=.98\linewidth]{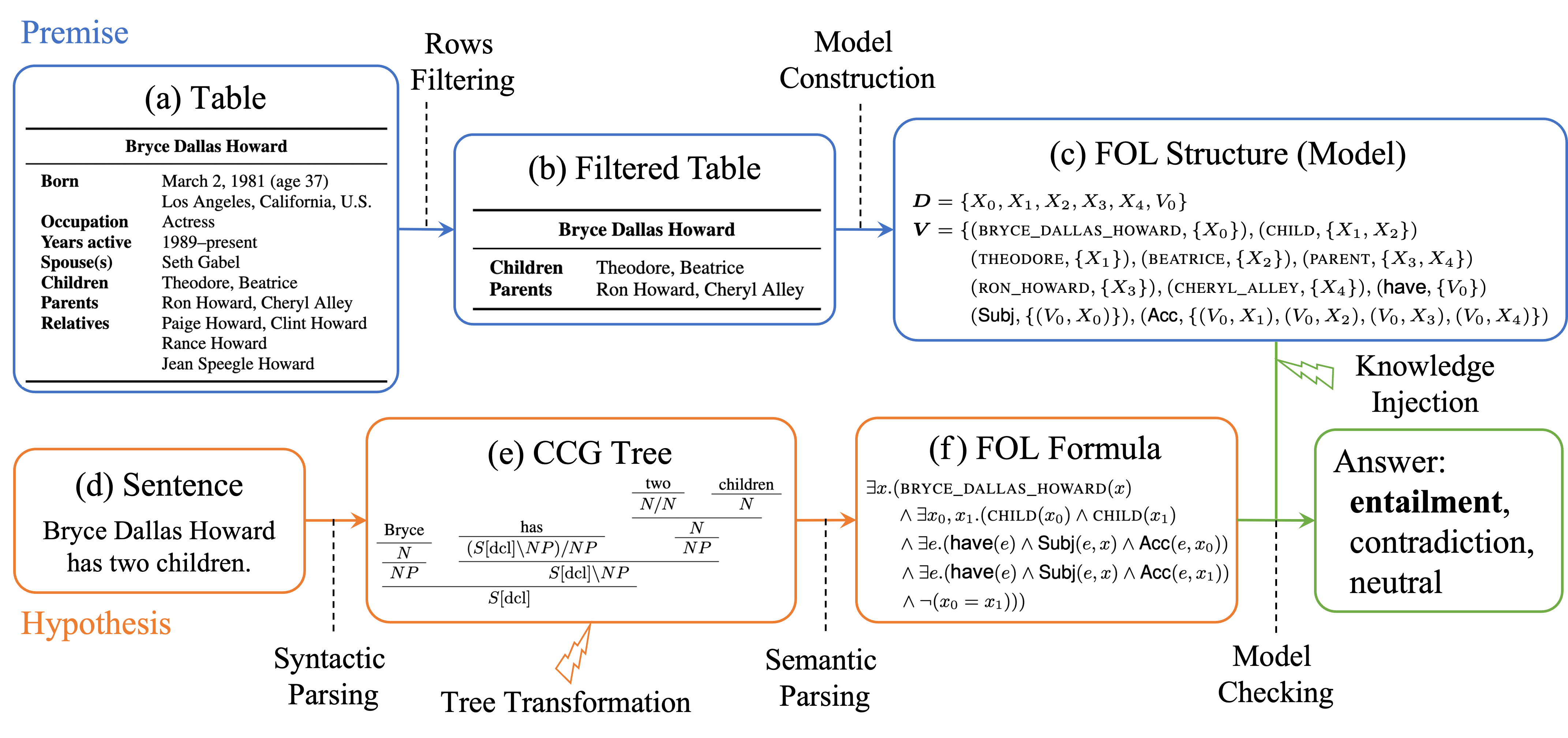}
    \caption{Overview of our proposed system with the example set for premise-hypothesis pair describing Bryce Dallas Howard.
    Our system returns \textit{true} (\textit{entailment}) for this pair.}
    \label{figure:method-overview}
\end{figure*}

\subsection{Model Checking}
\label{section:model-checking}
We use model checking in the Natural Language Toolkit (NLTK) \citep{bird-loper-2004-nltk,garrette-klein-2009-extensible} for making inference between tables and texts.
This system judges a truth-value of an FOL formula based on FOL structures.
An FOL structure (called \textit{model}) is defined by a pair of the domain $\bm{D}$ and the valuation $\bm{V}$, where $\bm{D}$ is a finite set of variables and $\bm{V}$ is a finite set of functions.
Each element of $\bm{V}$ is a pair of symbols, the name of the function and its domain. 

Based on the model used, the system will return
\begin{itemize}
    \setlength{\parskip}{-0.5mm}
    \item \textit{true} if the FOL formula is satisfiable,
    \item \textit{false} if the formula is unsatisfiable, and
    \item \textit{undefined} if there is an undefined function in the formula.
\end{itemize}
Figure \ref{figure:model-check-sample} shows outputs from model checking based on an example model and three formulas.

\section{Method}

\subsection{System Overview}
Figure \ref{figure:method-overview} shows the overview of our proposed system.
The system takes a table and a sentence as inputs and determines whether the table entails, contradicts, or is neutral toward the sentence.
We represent the meaning of tables as FOL structures (see Section \ref{section:meaning-representation-table}) and the meaning of sentences as FOL formulas (see Section \ref{section:meaning-representation-text}).
In the process of translating a table, we first make a filtered table, and then translate that table to an FOL structure. 

In the process of translating a sentence, we convert the sentence to a CCG derivation tree using a CCG parser \citep{yoshikawa-etal-2017-ccg}.
Before parsing, we use a Named Entity Recognition (NER) system in spaCy\footnote{\url{https://github.com/explosion/spaCy}} to identify a proper noun in sentences and add extra underscores to spaces and at the end of phrases so that such phrases can be categorized as one proper noun.
This derivation tree is modified by a tree transformation so that it handles numerical expressions correctly.
For the tree transformation, we use tsurgeon \citep{levy-andrew-2006-tregex} (see Appendix \ref{section:example-tt} for more details).
We then construct semantic representations (FOL formulas) of the hypothesis sentences according to the CCG derivation tree.
For semantic parsing based on CCG, we use ccg2lambda \citep{martinez-gomez-etal-2016-ccg2lambda}.
As a result, we obtain an FOL formula representing the whole sentence.

We apply model checking between the FOL structure and the FOL formula for inference using NLTK with optimization (see Section \ref{section:optimization}).
Under the FOL formula and the FOL structure, we assume
\begin{itemize}
    \setlength{\parskip}{-0.5mm}
    \item \textit{entailment} if our system returns \textit{true},
    \item \textit{contradiction} if our system returns \textit{false}, and
    \item \textit{neutral} otherwise.
\end{itemize}

\subsection{Meaning Representations for Tables}
\label{section:meaning-representation-table}
The top of the Figure \ref{figure:method-overview} shows the processes of translating from premise tables to FOL models.
We select the \textbf{Children} and \textbf{Parents} rows from the table (a) using rows filtering (see Section \ref{section:rows-filtering}).
Then, the filtered table (b) is translated into an FOL structure (c).
In (c), \textsf{have} is a meta-predicate (see Section \ref{section:model-construction}), a predicate connecting a title and other values.

\subsubsection{Rows Filtering} %
\label{section:rows-filtering}
To isolate rows from a premise table that are related to the hypothesis sentence, we apply Distracting Rows Removal (DRR), which was proposed by the previous approach \citep{neeraja-etal-2021-incorporating}.
Since that approach was NN-based, a sentence vector representation was generated for each row in the table, and the original DRR was applied to the sentence representation.
Then, the similarity score between each generated sentence and the hypothesis sentence was calculated.
In this process, the previous approach used fastText \citep{joulin2016fasttext} to obtain the embedding vectors of words.
They represented a hypothesis vector sequence of length $p$ as $(\bm{h}_0, \bm{h}_1, \ldots, \bm{h}_{p-1})$ and an $i$-th row vector sequence of length $q$ as $(\bm{t}^i_0, \bm{t}^i_1, \ldots, \bm{t}^i_{q-1})$.
The similarity score was then calculating using
\[
    \textsc{score}_i = \sum_{0\leq j<p}\max_{0\leq k<q} (\bm{h}_j\cdot \bm{t}^i_k)
\]
Finally, the four rows which were the most similar were selected as the premise.

We follow most of the original DRR, but with a slight modification.
First, since we directly represent a set of rows as FOL structures, we do not need to generate a sentence for each row.
Thus, our system makes a simple concatenation (not using any words) of keys and values rather than a proper sentence.
Also, to improve the similarity score calculation, we include numbers in a list of stopwords.
In rows filtering, we select the top two most similar rows as the premise.

\subsubsection{Model Construction}
\label{section:model-construction}
We construct a model based on the title and rows selected in Section \ref{section:rows-filtering}.
First, we define an entity variable $X_0$ that indicates a title.
For keys and values in rows, 
\begin{itemize}
    \item when the key is a noun, we define entity variables $X_i~(i\geq 1)$ indicating the value of each, and
    \item when the key is a verb, we define event variables $V_j~(j\geq 1)$, whose subject is the title entity and whose accusative is the value of each.
\end{itemize}
To classify the parts of speech of the keys as nouns or verbs of the keys, we use spaCy for part-of-speech (POS) tagging.
Keys are usually composed of nouns, verbs, adjectives, and prepositions, as shown in Figure \ref{figure:joe-biden}.
Since morphosyntactic ambiguity rarely appears in keys, we can classify keys into nouns and verbs by simply using a POS tagger.

We also introduce a meta-predicate \textsf{have}, with an event variable $V_0$.
The subject of \textsf{have} is the variable $X_0$ indicating the title entity, and the accusatives are any of the entities in values.

\begin{figure*}[t]
    \centering\tiny
    \input{figures/ccg_tree3.tex}
    \caption{A derivation tree of \textit{Bryce Dallas Howard has two children.} 
    \textsf{True} is a predicate which always returns true regardless of arity and argument.
    The function \textsc{bryce} is an abbreviation for \textsc{bryce\_dallas\_howard\_}.}
    \label{figure:ccg-tree}
\end{figure*}

\begin{table*}[t]
    \small\centering
    \begin{tabular}{lll} \toprule
    & Phrase & Logical formula \\ \midrule
(a)\!\!\!\! & less than two books\!\!\!\! & $\lambda F_1 F_2. \forall x_0 x_1. ((\book{x_0} \!\land\! \book{x_1} \land F_1(x_0) \land F_2(x_0)\land F_1(x_1)\land F_2(x_1))\!\to\! (x_0 \!=\! x_1))$ \\
(b)\!\!\!\! & at least two books\!\!\!\! & $\lambda F_1 F_2. \exists x_0 x_1. (\book{x_0} \land \book{x_1} \land F_1(x_0) \land F_2(x_0)\land F_1(x_1)\land F_2(x_1) \land \neg(x_0 \!=\! x_1))$ \\
(c)\!\!\!\! & twice & $\lambda V Q K. \exists e_1 e_2. (V(Q, \lambda e. (K(e)\land (e = e_1))) \land V(Q, \lambda e. (K(e)\land (e = e_2)))\land \neg(e_1 = e_2))$ \\ \bottomrule
\end{tabular}

    \caption{Examples of FOL formulas.
    $F_1$ and $F_2$ in (a) and (b) are unary predicates representing additional attributes of \textit{books} on the bottom of the syntactic tree. In (c), $V$ is a unary predicate for verb phrases, $Q$ is a binary predicate for noun phrases, and $K$ is a unary predicate for additional attributes of the event.} 
    \label{table:formula-ex}
\end{table*}

\subsubsection{Knowledge Injection} 
\label{section:knowledge-injection}
In some inference problems, an inference system needs to capture paraphrases (restatements of phrases that have the same meaning but are worded differently) in a premise table and a hypothesis sentence.
For example, the function \textsc{wife} is injected in a model because \textit{spouse} can be paraphrased as \textit{wife}.

Using knowledge graphs to paraphrase some words in keys, we calculate the relatedness score between each word in keys (\textit{key\_term}) and each word in the hypothesis sentence (\textit{hypo\_term}).
When the score exceeds the threshold (0.5), the \textit{hypo\_term} is introduced as a function, and the domain of which is the same as that of the \textit{key\_term}.
In this process, we use the standard knowledge graph ConceptNet \citep{liu-singh-2004-conceptnet} to get the relatedness score between
\textit{key\_term} and \textit{hypo\_term}.
ConceptNet is a knowledge base that includes WordNet~\citep{miller-1995-wordnet}.
We select ConceptNet because InfoTabS requires paraphrases based on not only hypernymy and hyponymy relations considered in WordNet, but also common knowledge.
For example, to understand whether the hypothesis \textit{Joe Biden has married twice} is entailed or not by Figure \ref{figure:joe-biden}, we need to capture paraphrases between \textbf{Spouse} in the premise table and \textit{marry} or \textit{marriage} in the hypothesis.

\subsection{Meaning Representations for Sentences}
\label{section:meaning-representation-text}
We construct meaning representations of hypothesis sentences based on the CCG derivation tree and Neo-Davidsonian Event Semantics \citep{Parsons1990-PAREIT}.
ccg2lambda \citep{mineshima-etal-2015-higher,martinez-gomez-etal-2016-ccg2lambda} is used to obtain meaning representations (FOL formulas) of hypothesis sentences based on CCG and $\lambda$-calculus.
We extend the semantic template that defines lexical entries and schematic entries assigned to CCG categories in \citet{mineshima-etal-2015-higher} so that it can handle the numerical expressions for this task. 
In total, we add 251 extra lexical entries for the numerical expressions.
Figure \ref{figure:ccg-tree} shows an example of CCG derivation trees with meaning representations involving numerical expressions. %

We focus on expressions related to numerical comparatives: \textit{less than}, \textit{no more than}, \textit{exactly}, \textit{at least}, \textit{no less than}, and \textit{more than}.
We need to consider how to represent the meaning of a noun phrase ($NP$ as its CCG category) that involves a numerical comparative and the number of entities, such as \textit{less than two books}.
The meaning of this phrase is analyzed in Table \ref{table:formula-ex}a.
We also analyze the meaning of the phrase \textit{at least two books} in Table \ref{table:formula-ex}b.
The meaning representation of \textit{exactly two books} is given as the composition of the representation of \textit{at least two books} and the representation of \textit{no more than two books} \citep{van-benthem-1986-logical}.

Adverbs of frequency such as \textit{twice} describe the number of events, and their CCG category is $(S\back NP)\back (S\back NP)$.
The semantic representation of \textit{twice} is given in Table \ref{table:formula-ex}c.

In previous work, \citet{haruta-etal-2020-combining,haruta-etal-2020-logical} handled generalized quantifiers including numerical comparatives as binary predicates \textsf{many}.
For example, the noun phrase \textit{two cats} is represented as $\textsc{cat}(x)\wedge\textsf{many}(x, 2)$, which indicates that $x$ has the property of \textsc{cat} and is composed of at least 2 entities.
Since one of the aims of our system is to count the elements in the values of premise tables, our system assigns different entities for every word or phrase in the values.

\subsection{Optimization of Model Checking}
\label{section:optimization}
To optimize the process of model checking between tables and texts, we extend the implementation of model checking in NLTK.
Figure \ref{source:eval} shows the program that evaluates the truth-value of $\exists x. A$.
NLTK is implemented in Python and uses a set, which is an unordered collection, to represent the  domain $\bm{D}$ of an FOL structure.
When evaluating a for loop with a set (line 1 of Figure \ref{source:eval}), an order of values in the set is not fixed for each run. 
To fix the order, we changed the implementation of the domain from a set to a list.

\begin{figure}[t]
    \begin{algorithm}[H]
        \small
        \begin{algorithmic}[1]
            \FOR{$y$ in $\bm{D}$}
                \IF{the truth-value of $A[y/x]$ is true}
                    \RETURN\TRUE
                \ENDIF
            \ENDFOR 
            \RETURN \FALSE
        \end{algorithmic}
    \end{algorithm}
    \caption{A program for evaluating the truth-value of $\exists x. A$.}
    \label{source:eval}
\end{figure}

We also modify the original program for model checking in NLTK to make judgments faster. 
First, we sort the domain $\bm{D}$ to facilitate faster evaluation,  giving 
$(X_0, X_1, \ldots, X_{n-1}, V_0, V_1, \ldots, V_{m-1})$,
where $n$ and $m$ are the number of entities and events, respectively.
It is sorted this way because the title variable $X_0$ is often the subject of the hypothesis sentence, which can be found at the top of the meaning representations.

Second, we use constraints for both the existential and universal quantifiers $(\exists ~\textrm{and}~ \forall)$. 
We do not substitute one variable for the other type of bounded variable in the evaluation scheme during quantification. 
Third, we use constraints for existential quantifiers $(\exists)$ so as not to use the same variables for two or more bounded variables during substitution.
We apply this restriction for only to entity variables because the same variable may be applied to different bounded variables for each event.
In the process of model checking, we set a timeout of 10 seconds for judging whether the formula is satisfiable.

\section{Experiments} 
We evaluate the extent to which our system can perform inference with numerical comparatives.
We make an evaluation protocol that focuses on the numerical understanding between semi-structured tables and texts in English.

\begin{table}[t]
    \small\centering
    \begin{tabular}{ll} \toprule
    \multicolumn{2}{c}{\textbf{Karachi}} \\ \midrule
    \textbf{Country} & Pakistan \\
    \textbf{Province} & Sindh \\ %
    \textbf{Metropolitan} & 2011 \\ 
    \textbf{corporation} &  \\ %
    \textbf{City council} & City Complex, Gulshan-e-Iqbal Town\\ %
    \textbf{Districts} & Central Karachi, East Karachi, South  \\
    & Karachi, West Karachi, Korangi, Malir \\ \bottomrule
\end{tabular}
    \caption{The premise table for the hypothesis \textit{Karachi has a half dozen districts.}}
    \label{table:karachi}
\end{table}

\subsection{Dataset} %
\label{section:dataset}
We created a new dataset for the numerical understanding of semi-structured tables.
There are two motivations for doing so. 
One is that the number of test cases for numerical understanding is limited to the previous NLI dataset for semi-structured tables, InfoTabS \citep{gupta-etal-2020-infotabs}. 
In addition, to evaluate whether NLI systems consistently perform inference with numerical comparatives, we need to analyze whether the prediction labels change correctly when the numbers in the hypothesis sentence are slightly changed from those in the original hypothesis sentence.

To create the dataset for numerical understanding of semi-structured tables, we first manually extracted 105 examples involving numerical expressions from the $\alpha_1, \alpha_2$, and $\alpha_3$ test sets in InfoTabS.
The inference for these examples requires an understanding of the number of entities and events.
We then made a \textit{problem set} from each example and defined the \textit{base hypothesis} of the test cases by rewriting to the actual value $n$ with \textit{exactly} entailed from a premise table.

Table \ref{table:karachi} shows a premise table for the hypothesis \textit{Karachi has a half dozen districts}, which was extracted from InfoTabS. 
This premise-hypothesis pair is an example, and it makes a problem set for the statement \textit{how many districts Karachi has}.
Because we can precisely see six districts in Karachi from the premise table,
the base hypothesis of this problem set is \textit{Karachi has exactly six districts}, where \textit{a half dozen} is defined as the number \textit{six}.
When the gold label of an example extracted from InfoTabS is \textit{neutral}, a base hypothesis of the example is made by simply replacing the numerical comparatives with \textit{exactly}.
The gold label of the base hypothesis is the same as that of the original example.
For instance, if the original hypothesis is \textit{Bob has more than two dogs}, and its gold label is \textit{neutral}, then the base hypothesis becomes \textit{Bob has exactly two dogs}. 
Finally, we make test cases from each base hypothesis using the following process:
\begin{enumerate}[(i)]
    \item We make a new hypothesis sentence $S$ by removing \textit{exactly} from the base hypothesis.
    \item We make two new hypothesis sentences, $S_+$ and $S_-$ by replacing the number $n$ in $S$ with $n+1$ and $n-1$ in $S$, respectively.
    \item We make six additional hypothesis sentences each from $S, S_+$, and $S_-$ by adding the expressions related to numerical comparatives, \textit{less than}, \textit{no more than}, \textit{exactly}, \textit{at least}, \textit{no less than}, and \textit{more than}, thus making a problem set consisting of 21 hypothesis sentences with correct gold labels. 
    Table \ref{table:karachi-testcases} shows a part of the hypothesis sentences.
    \item We remove unnatural hypothesis sentences from the problem set, including such as \textit{at least zero} and \textit{less than one}.
\end{enumerate}
Note that here \textit{two} has the same meaning as \textit{at least two}. 
Our dataset consists of 105 problem sets with 1,979 test cases.
The distribution of gold labels is (\textit{entailment}, \textit{neutral}, \textit{contradiction})${} = {}$(965, 176, 838).
This dataset includes ten problem sets that are filled with \textit{neutral} labels.
We confirmed all words are commonly used in a training set in InfoTabS and our dataset.

\begin{table}[t]
    \small\centering
    \begin{tabular}{lcc} \toprule
    Hypothesis                                & Gold    & Note      \\ \midrule
    Karachi has less than five districts.     & C & $[2]$ \\
    Karachi has less than six districts.      & C & $[1]$ \\
    Karachi has less than seven districts.    & E    &           \\ \midrule
    Karachi has five districts.               & E    & $[1]$ \\
    Karachi has six districts.                & E    &           \\
    Karachi has seven districts.              & C &           \\ \midrule
    Karachi has more than five districts.     & E    & $[1]$ \\
    Karachi has more than six districts.      & C &           \\
    Karachi has more than seven districts.    & C &           \\ \bottomrule
\end{tabular}
    \caption{A part of the test cases made from the problem set for the base hypothesis \textit{Karachi has exactly six districts}.
    $[i] ~(i = 1, 2)$ as noted means that the test case is not defined when $n \leq  i$, $n$ being the actual value.
    E and C are \textit{entailment} and \textit{contradiction}, respectively.}
    \label{table:karachi-testcases}
\end{table}

\subsection{Experimental Setup for Previous Research}
\label{section:setup-previous}
\citet{neeraja-etal-2021-incorporating} proposed an NN-based model for inference between semi-structured tables and texts and tested it by InfoTabS.
We compare our system to +KG explicit, which was the setting for which the previous model \citep{neeraja-etal-2021-incorporating} achieved the highest performance.
+KG explicit consists of the following four methods for making sentence representations of tables.
\paragraph{Implicit Knowledge Addition} The model adds information that is not in the tables and texts to models by pre-training with a large-scale NLI corpus, MultiNLI \citep{williams-etal-2018-broad}.
\paragraph{Better Paragraph Representation} The model generates more grammatical sentences for specific entity types, such as money, date, and cardinal, with carefully crafted templates when making sentence representations of tables.
\paragraph{Distracting Rows Removal (DRR)} The model removes several rows from the premise table that are unrelated to the hypothesis sentence.
For a detailed explanation of DRR, see Section \ref{section:rows-filtering}.
\paragraph{Explicit Knowledge Addition} The model adds a suitable meaning to the keys for each premise from WordNet \citep{miller-1995-wordnet} or Wikipedia articles by calculating similarity based on the BERT embedding.

+KG explicit makes sentence representations of tables and uses RoBERTa-large \citep{liu2019roberta} for encoding premise-hypothesis pairs.
Almost all of the setups are identical to what was used in previous research except 
(i) the batch size is set to 4 and
(ii) we adopt the result of one seed rather than the average of three seeds.

\begin{table}[t]
    \small\centering
    \begin{tabular}{lcc} \toprule
& +KG  & Ours \\ \midrule
All problem sets & 0.03 & \textbf{0.31} \\
All problem sets excluding \textit{neutral}-filled & 0.00 & \textbf{0.27} \\\bottomrule
\end{tabular}
    \caption{The accuracy of problem sets whose test cases were all predicted correctly.
    +KG is an abbreviation for +KG explicit.
    }
    \label{table:result-accuracy}
\end{table}

\subsection{Results} 
\paragraph{Accuracy per Problem Set}
Table \ref{table:result-accuracy} shows the accuracy of the previous model (+KG) and our system (Ours) for a number of problem sets.
Our proposed system could correctly predict 31\% of all problem sets, while the previous model only predicted 3\%. 
Premise-hypothesis pairs whose gold labels are \textit{neutral} can be predicted correctly without a precise numerical understanding.
Table \ref{table:result-accuracy} also shows that +KG could not perform inference on any problem set whose gold labels were \textit{entailment} or \textit{contradiction} at all.
On the other hand, the accuracy of our logic-based system was 27\%. 
These results indicate that our system better handles inference involving numerical comparatives than the previous model, being able to more robustly predict \textit{entailment} and \textit{contradiction} labels.
This shows that our proposed dataset for numerical understanding is challenging for current systems.
We describe the error analysis of our system in the fourth paragraph of this section.

\begin{table}[t]
    \small\centering
    \begin{tabular}{lcc} \toprule
             & +KG  & Ours \\ \midrule
less than $k$   & 0.10 & \textbf{0.36} \\
no more than $k$ & 0.10 & \textbf{0.35} \\
exactly $k$     & 0.19 & \textbf{0.32} \\
$k$      & 0.24 & \textbf{0.33} \\
at least $k$    & 0.08 & \textbf{0.32} \\
no less than $k$ & 0.19 & \textbf{0.33} \\
more than $k$   & 0.17 & \textbf{0.35} \\\bottomrule
\end{tabular}
    \caption{The accuracy for each numerical comparative construction. 
    +KG is an abbreviation for +KG explicit.
    $k$ indicates a number.}
    \label{table:result-quantifier}
\end{table}

\paragraph{Understanding for Each Numerical Comparative}
Table \ref{table:result-quantifier} shows the accuracy of both methods for each numerical comparative construction.
We observe that our proposed method can predict correct labels more often than the existing method for all numerical comparatives. 

\paragraph{Run Time for Model Checking with Optimization}
We compare the run times for model checking with and without our optimization for model checking (see Section \ref{section:optimization}).
We chose six problem sets involving different numbers of values, which
consist of two problem sets each whose numbers of values are 2, 4, and 6. 
All of the problems require understanding the number of entities.
The number of test cases is 124.
Table \ref{table:result-modelcheck} shows the average and maximum run times for ten trials.
We observe that our optimization made model checking much faster. %

\begin{figure*}[t]
    \fbox{
    \begin{minipage}[t]{.47\hsize}
        \small\centering
        \begin{tabular}{ll}\toprule
    \multicolumn{2}{c}{\textbf{Jodie Whittaker}} \\ \midrule
    \textbf{Spouse} & Christian Contreras \\\bottomrule
\end{tabular}

        \subcaption*{i. Part of the filtered table describing Jodie Whittaker.}

        \vspace{-1\baselineskip}
        \input{figures/jodie_model.tex}
        \vspace{-1.5\baselineskip}
        \subcaption*{ii. Part of the model constructed by our system for (a-i).}

        \vspace{-1.5\baselineskip}
        \begin{multline*}
            \exists x. (\textsc{jodie\_whittaker\_}(x)\land \exists x_0. (\underline{\textsc{husband}(x_0)}\\
            \land \exists e. (\textsf{have}(e)\land \textsf{Subj}(e, x)\land \textsf{Acc}(e, x_0))))
        \end{multline*}
        \vspace{-1.5\baselineskip}
        \subcaption*{iii. An FOL formula constructed from the hypothesis \textit{Jodie Whittaker has had one husband}.}

        \caption{Outputs of our system to the premise-hypothesis pair describing Jodie Whittaker.
        Our system was able to perform inference correctly.}
        \label{figure:jodie-all}
    \end{minipage} 
    }
    \fbox{
    \begin{minipage}[t]{.47\hsize}
        \small\centering
        \begin{tabular}{ll}\toprule
    \multicolumn{2}{c}{\textbf{Karl Ferdinand Braun}} \\ \midrule
    \textbf{Awards} & Nobel Prize in Physics (1909) \\\bottomrule
\end{tabular}

        \subcaption*{i. Part of the filtered table describing Karl Ferdinand Braun.}

        \vspace{-1\baselineskip}
        \input{figures/karl_model.tex}
        \vspace{-1\baselineskip}
        \subcaption*{ii. Part of the model constructed by our system for (b-i).}

        \vspace{-1\baselineskip}
        \begin{multline*}
            \exists x. (\textsc{karl}(x)\land \exists x_0. (\textsc{award}(x_0)\\
            \land \exists e. (\underline{\textsc{win}(e)}\land \textsf{Subj}(e, x)\land \textsf{Acc}(e, x_0))))
        \end{multline*}
        \vspace{-1.5\baselineskip}
        \subcaption*{iii. An FOL formula constructed from the hypothesis \textit{Karl Ferdinand Braun won one award}.}

        \caption{Outputs of our system to the premise-hypothesis pair describing Karl Ferdinand Braun.
        Our system was not able to perform inference correctly.}
        \label{figure:karl-all}
    \end{minipage}
    }

    \caption{Two premise-hypothesis pairs, one for which our system was able to perform inference (\subref{figure:jodie-all}) and one for which it was not (\subref{figure:karl-all}).
    The function \textsc{karl} in (\subref{figure:karl-all}-ii, \subref{figure:karl-all}-iii) is an abbreviation for \textsc{karl\_ferdinand\_braun\_}.
    The underlined functions are added in the knowledge injection process to perform inference.}
    \label{figure:error-verb}
\end{figure*}

\paragraph{Error Analysis}
\label{section:error-analysis}
Error analysis shows that main errors are caused by the failure of knowledge injection.
Figure \ref{figure:error-verb} shows two premise-hypothesis pairs, one for which our system was able to perform inference and one for which it was not.
In Figure \ref{figure:jodie-all}, the function \textsc{husband} was added to the model in the knowledge injection process because the relatedness score between \textit{spouse} and \textit{husband} was high (0.747).
On the other hand, in Figure \ref{figure:karl-all}, the function \textsc{win} was not added to the model because the relatedness score between \textit{award} and \textit{win} was low (0.336).
In addition, even though we improved the speed of the original model checking program, several test cases still ran out of time. 
For example, the problem with the hypothesis sentence \textit{Jimmy Eat World has been on 13 labels} (this gold label is \textit{contradiction}) exceeded the maximum time limit (10 seconds).

\begin{table}[t]
    \small\centering
    
\begin{tabular}{lrr} \toprule
Optimization & Average & Maximum \\ \midrule
disabled     & 3.20 & 185.17 \\
enabled      & \textbf{0.04} & \textbf{1.26} \\ \bottomrule
\end{tabular}

    \caption{Average and maximum run time (seconds) for model checking with and without optimization.}
    \label{table:result-modelcheck}
\end{table}

\paragraph{Discussion}
We discuss how to handle various types of inference other than the numerical one in InfoTabS with our inference system. 
First, we have to correctly parse values in various tables and extract information from them.
For example, to determine whether Hypothesis 1 in Figure \ref{figure:joe-biden} is entailed by the premise table, 
we need to parse the noun phrase \textit{November 20, 1942} into one date format.
In addition to this, various formats are needed to be provided, such as age, duration, and year of marriage.
Also, some test cases require arithmetic operations other than counting, such as \textit{Joe Biden and Neilia Hunter divorced six years after their marriage}, based on the premise table in Figure \ref{figure:joe-biden}.
Although such issues are tricky, we believe that our logic-based approach is applicable with adding premises related to arithmetic operations.

\section{Conclusion}
In this study, we proposed a logic-based system for an NLI task that requires numerical understanding in semi-structured tables.
We built an NLI dataset that focuses on numerical comparatives between semi-structured tables and texts.
Using this dataset, we showed that our system performed more robustly than the previous NN-based model.

In future work, we will improve knowledge injection process to cover various problems. We also seek to handle other generalized quantifiers such as \textit{many}.
We believe that our system and dataset for performing numerical inference between semi-structured tables and texts could pave the way for applications of inference between resources other than texts.

\section*{Acknowledgements}
We thank the three anonymous reviewers for their helpful comments and feedback.
This work was supported by PRESTO, JST Grant Number JPMJPR21C8, Japan.

\bibliography{anthology,custom}

\begin{thebibliography}{34}
\expandafter\ifx\csname natexlab\endcsname\relax\def\natexlab#1{#1}\fi

\bibitem[{Abzianidze(2015)}]{abzianidze-2015-tableau}
Lasha Abzianidze. 2015.
\newblock \href {https://doi.org/10.18653/v1/D15-1296} {A tableau prover for
  natural logic and language}.
\newblock In \emph{Proceedings of the 2015 Conference on Empirical Methods in
  Natural Language Processing}, pages 2492--2502, Lisbon, Portugal. Association
  for Computational Linguistics.

\bibitem[{Bird and Loper(2004)}]{bird-loper-2004-nltk}
Steven Bird and Edward Loper. 2004.
\newblock \href {https://aclanthology.org/P04-3031} {{NLTK}: The natural
  language toolkit}.
\newblock In \emph{Proceedings of the {ACL} Interactive Poster and
  Demonstration Sessions}, pages 214--217, Barcelona, Spain. Association for
  Computational Linguistics.

\bibitem[{Bos(2008)}]{bos-2008-wide}
Johan Bos. 2008.
\newblock \href {https://aclanthology.org/W08-2222} {Wide-coverage semantic
  analysis with {B}oxer}.
\newblock In \emph{Semantics in Text Processing. {STEP} 2008 Conference
  Proceedings}, pages 277--286. College Publications.

\bibitem[{Clark and Curran(2007)}]{clark-curran-2007-wide}
Stephen Clark and James~R. Curran. 2007.
\newblock \href {https://doi.org/10.1162/coli.2007.33.4.493} {Wide-coverage
  efficient statistical parsing with {CCG} and log-linear models}.
\newblock \emph{Computational Linguistics}, 33(4):493--552.

\bibitem[{Dagan et~al.(2006)Dagan, Glickman, and Magnini}]{dagan-etal-2005-rte}
Ido Dagan, Oren Glickman, and Bernardo Magnini. 2006.
\newblock The pascal recognising textual entailment challenge.
\newblock In \emph{Machine Learning Challenges. Evaluating Predictive
  Uncertainty, Visual Object Classification, and Recognising Tectual
  Entailment}, pages 177--190, Berlin, Heidelberg. Springer Berlin Heidelberg.

\bibitem[{Devlin et~al.(2019)Devlin, Chang, Lee, and
  Toutanova}]{devlin-etal-2019-bert}
Jacob Devlin, Ming-Wei Chang, Kenton Lee, and Kristina Toutanova. 2019.
\newblock \href {https://doi.org/10.18653/v1/N19-1423} {{BERT}: Pre-training of
  deep bidirectional transformers for language understanding}.
\newblock In \emph{Proceedings of the 2019 Conference of the North {A}merican
  Chapter of the Association for Computational Linguistics: Human Language
  Technologies, Volume 1 (Long and Short Papers)}, pages 4171--4186,
  Minneapolis, Minnesota. Association for Computational Linguistics.

\bibitem[{Garrette and Klein(2009)}]{garrette-klein-2009-extensible}
Dan Garrette and Ewan Klein. 2009.
\newblock \href {https://aclanthology.org/W09-3712} {An extensible toolkit for
  computational semantics}.
\newblock In \emph{Proceedings of the Eight International Conference on
  Computational Semantics}, pages 116--127, Tilburg, The Netherlands.
  Association for Computational Linguistics.

\bibitem[{Gupta et~al.(2020)Gupta, Mehta, Nokhiz, and
  Srikumar}]{gupta-etal-2020-infotabs}
Vivek Gupta, Maitrey Mehta, Pegah Nokhiz, and Vivek Srikumar. 2020.
\newblock \href {https://doi.org/10.18653/v1/2020.acl-main.210} {{INFOTABS}:
  Inference on tables as semi-structured data}.
\newblock In \emph{Proceedings of the 58th Annual Meeting of the Association
  for Computational Linguistics}, pages 2309--2324, Online. Association for
  Computational Linguistics.

\bibitem[{Haruta et~al.(2020{\natexlab{a}})Haruta, Mineshima, and
  Bekki}]{haruta-etal-2020-combining}
Izumi Haruta, Koji Mineshima, and Daisuke Bekki. 2020{\natexlab{a}}.
\newblock \href {https://doi.org/10.18653/v1/2020.coling-main.156} {Combining
  event semantics and degree semantics for natural language inference}.
\newblock In \emph{Proceedings of the 28th International Conference on
  Computational Linguistics}, pages 1758--1764, Barcelona, Spain (Online).
  International Committee on Computational Linguistics.

\bibitem[{Haruta et~al.(2020{\natexlab{b}})Haruta, Mineshima, and
  Bekki}]{haruta-etal-2020-logical}
Izumi Haruta, Koji Mineshima, and Daisuke Bekki. 2020{\natexlab{b}}.
\newblock \href {https://doi.org/10.18653/v1/2020.acl-srw.35} {Logical
  inferences with comparatives and generalized quantifiers}.
\newblock In \emph{Proceedings of the 58th Annual Meeting of the Association
  for Computational Linguistics: Student Research Workshop}, pages 263--270,
  Online. Association for Computational Linguistics.

\bibitem[{Hu et~al.(2020)Hu, Chen, Richardson, Mukherjee, Moss, and
  Kuebler}]{hu-etal-2020-monalog}
Hai Hu, Qi~Chen, Kyle Richardson, Atreyee Mukherjee, Lawrence~S. Moss, and
  Sandra Kuebler. 2020.
\newblock \href {https://aclanthology.org/2020.scil-1.40} {{M}ona{L}og: a
  lightweight system for natural language inference based on monotonicity}.
\newblock In \emph{Proceedings of the Society for Computation in Linguistics
  2020}, pages 334--344, New York, New York. Association for Computational
  Linguistics.

\bibitem[{Jia and Liang(2017)}]{jia-liang-2017-adversarial}
Robin Jia and Percy Liang. 2017.
\newblock \href {https://doi.org/10.18653/v1/D17-1215} {Adversarial examples
  for evaluating reading comprehension systems}.
\newblock In \emph{Proceedings of the 2017 Conference on Empirical Methods in
  Natural Language Processing}, pages 2021--2031, Copenhagen, Denmark.
  Association for Computational Linguistics.

\bibitem[{Joulin et~al.(2016)Joulin, Grave, Bojanowski, Douze, J{\'e}gou, and
  Mikolov}]{joulin2016fasttext}
Armand Joulin, Edouard Grave, Piotr Bojanowski, Matthijs Douze, H{\'e}rve
  J{\'e}gou, and Tomas Mikolov. 2016.
\newblock \href {http://arxiv.org/abs/1612.03651} {{F}ast{T}ext.zip:
  Compressing text classification models}.
\newblock \emph{Computing Research Repository}, arXiv:1612.03651.

\bibitem[{Krishnamurthy et~al.(2017)Krishnamurthy, Dasigi, and
  Gardner}]{krishnamurthy-etal-2017-neural}
Jayant Krishnamurthy, Pradeep Dasigi, and Matt Gardner. 2017.
\newblock \href {https://doi.org/10.18653/v1/D17-1160} {Neural semantic parsing
  with type constraints for semi-structured tables}.
\newblock In \emph{Proceedings of the 2017 Conference on Empirical Methods in
  Natural Language Processing}, pages 1516--1526, Copenhagen, Denmark.
  Association for Computational Linguistics.

\bibitem[{Lebret et~al.(2016)Lebret, Grangier, and
  Auli}]{lebret-etal-2016-neural}
R{\'e}mi Lebret, David Grangier, and Michael Auli. 2016.
\newblock \href {https://doi.org/10.18653/v1/D16-1128} {Neural text generation
  from structured data with application to the biography domain}.
\newblock In \emph{Proceedings of the 2016 Conference on Empirical Methods in
  Natural Language Processing}, pages 1203--1213, Austin, Texas. Association
  for Computational Linguistics.

\bibitem[{Levy and Andrew(2006)}]{levy-andrew-2006-tregex}
Roger Levy and Galen Andrew. 2006.
\newblock \href {http://www.lrec-conf.org/proceedings/lrec2006/pdf/513_pdf.pdf}
  {Tregex and tsurgeon: tools for querying and manipulating tree data
  structures}.
\newblock In \emph{Proceedings of the Fifth International Conference on
  Language Resources and Evaluation ({LREC}{'}06)}, Genoa, Italy. European
  Language Resources Association (ELRA).

\bibitem[{Liu and Singh(2004)}]{liu-singh-2004-conceptnet}
Hugo Liu and Push Singh. 2004.
\newblock Conceptnet - a practical commonsense reasoning tool-kit.
\newblock \emph{BT Technology Journal}, 22:211--226.

\bibitem[{Liu et~al.(2019)Liu, Ott, Goyal, Du, Joshi, Chen, Levy, Lewis,
  Zettlemoyer, and Stoyanov}]{liu2019roberta}
Yinhan Liu, Myle Ott, Naman Goyal, Jingfei Du, Mandar Joshi, Danqi Chen, Omer
  Levy, Mike Lewis, Luke Zettlemoyer, and Veselin Stoyanov. 2019.
\newblock \href {http://arxiv.org/abs/1907.11692} {{R}o{BERT}a: A robustly
  optimized bert pretraining approach}.
\newblock \emph{Computing Research Repository}, arXiv:1907.11692.

\bibitem[{Mart{\'\i}nez-G{\'o}mez et~al.(2016)Mart{\'\i}nez-G{\'o}mez,
  Mineshima, Miyao, and Bekki}]{martinez-gomez-etal-2016-ccg2lambda}
Pascual Mart{\'\i}nez-G{\'o}mez, Koji Mineshima, Yusuke Miyao, and Daisuke
  Bekki. 2016.
\newblock \href {https://doi.org/10.18653/v1/P16-4015} {ccg2lambda: A
  compositional semantics system}.
\newblock In \emph{Proceedings of {ACL}-2016 System Demonstrations}, pages
  85--90, Berlin, Germany. Association for Computational Linguistics.

\bibitem[{Miller(1995)}]{miller-1995-wordnet}
George~A. Miller. 1995.
\newblock \href {https://doi.org/10.1145/219717.219748} {Wordnet: A lexical
  database for english}.
\newblock \emph{Commun. ACM}, 38(11):39--41.

\bibitem[{Mineshima et~al.(2015)Mineshima, Mart{\'\i}nez-G{\'o}mez, Miyao, and
  Bekki}]{mineshima-etal-2015-higher}
Koji Mineshima, Pascual Mart{\'\i}nez-G{\'o}mez, Yusuke Miyao, and Daisuke
  Bekki. 2015.
\newblock \href {https://doi.org/10.18653/v1/D15-1244} {Higher-order logical
  inference with compositional semantics}.
\newblock In \emph{Proceedings of the 2015 Conference on Empirical Methods in
  Natural Language Processing}, pages 2055--2061, Lisbon, Portugal. Association
  for Computational Linguistics.

\bibitem[{Naik et~al.(2018)Naik, Ravichander, Sadeh, Rose, and
  Neubig}]{naik-etal-2018-stress}
Aakanksha Naik, Abhilasha Ravichander, Norman Sadeh, Carolyn Rose, and Graham
  Neubig. 2018.
\newblock \href {https://aclanthology.org/C18-1198} {Stress test evaluation for
  natural language inference}.
\newblock In \emph{Proceedings of the 27th International Conference on
  Computational Linguistics}, pages 2340--2353, Santa Fe, New Mexico, USA.
  Association for Computational Linguistics.

\bibitem[{Neeraja et~al.(2021)Neeraja, Gupta, and
  Srikumar}]{neeraja-etal-2021-incorporating}
J.~Neeraja, Vivek Gupta, and Vivek Srikumar. 2021.
\newblock \href {https://doi.org/10.18653/v1/2021.naacl-main.224}
  {Incorporating external knowledge to enhance tabular reasoning}.
\newblock In \emph{Proceedings of the 2021 Conference of the North American
  Chapter of the Association for Computational Linguistics: Human Language
  Technologies}, pages 2799--2809, Online. Association for Computational
  Linguistics.

\bibitem[{Parsons(1990)}]{Parsons1990-PAREIT}
Terence Parsons. 1990.
\newblock \emph{Events in the Semantics of English: A Study in Subatomic
  Semantics}.
\newblock The MIT Press, Cambridge, MA.

\bibitem[{Pasupat and Liang(2015)}]{pasupat-liang-2015-compositional}
Panupong Pasupat and Percy Liang. 2015.
\newblock \href {https://doi.org/10.3115/v1/P15-1142} {Compositional semantic
  parsing on semi-structured tables}.
\newblock In \emph{Proceedings of the 53rd Annual Meeting of the Association
  for Computational Linguistics and the 7th International Joint Conference on
  Natural Language Processing (Volume 1: Long Papers)}, pages 1470--1480,
  Beijing, China. Association for Computational Linguistics.

\bibitem[{Ravichander et~al.(2019)Ravichander, Naik, Rose, and
  Hovy}]{ravichander-etal-2019-equate}
Abhilasha Ravichander, Aakanksha Naik, Carolyn Rose, and Eduard Hovy. 2019.
\newblock \href {https://doi.org/10.18653/v1/K19-1033} {{EQUATE}: A benchmark
  evaluation framework for quantitative reasoning in natural language
  inference}.
\newblock In \emph{Proceedings of the 23rd Conference on Computational Natural
  Language Learning (CoNLL)}, pages 349--361, Hong Kong, China. Association for
  Computational Linguistics.

\bibitem[{Richardson et~al.(2020)Richardson, Hu, Moss, and
  Sabharwal}]{Richardson_Hu_Moss_Sabharwal_2020}
Kyle Richardson, Hai Hu, Lawrence Moss, and Ashish Sabharwal. 2020.
\newblock \href {https://doi.org/10.1609/aaai.v34i05.6397} {Probing natural
  language inference models through semantic fragments}.
\newblock \emph{Proceedings of the AAAI Conference on Artificial Intelligence},
  34(05):8713--8721.

\bibitem[{Rozen et~al.(2019)Rozen, Shwartz, Aharoni, and
  Dagan}]{rozen-etal-2019-diversify}
Ohad Rozen, Vered Shwartz, Roee Aharoni, and Ido Dagan. 2019.
\newblock \href {https://doi.org/10.18653/v1/K19-1019} {Diversify your
  datasets: Analyzing generalization via controlled variance in adversarial
  datasets}.
\newblock In \emph{Proceedings of the 23rd Conference on Computational Natural
  Language Learning (CoNLL)}, pages 196--205, Hong Kong, China. Association for
  Computational Linguistics.

\bibitem[{Steedman(2000)}]{steedman-2000-ccg}
Mark Steedman. 2000.
\newblock \emph{The Syntactic Process}.
\newblock The MIT Press, Cambridge, MA.

\bibitem[{Suzuki et~al.(2019)Suzuki, Yanaka, Yoshikawa, Mineshima, and
  Bekki}]{suzuki-etal-2019-multimodal}
Riko Suzuki, Hitomi Yanaka, Masashi Yoshikawa, Koji Mineshima, and Daisuke
  Bekki. 2019.
\newblock \href {https://doi.org/10.18653/v1/P19-2054} {Multimodal logical
  inference system for visual-textual entailment}.
\newblock In \emph{Proceedings of the 57th Annual Meeting of the Association
  for Computational Linguistics: Student Research Workshop}, pages 386--392,
  Florence, Italy. Association for Computational Linguistics.

\bibitem[{van Benthem(1986)}]{van-benthem-1986-logical}
Johan van Benthem. 1986.
\newblock \emph{Essays in Logical Semantics}.
\newblock Springer, Dordrecht.

\bibitem[{Williams et~al.(2018)Williams, Nangia, and
  Bowman}]{williams-etal-2018-broad}
Adina Williams, Nikita Nangia, and Samuel Bowman. 2018.
\newblock \href {https://doi.org/10.18653/v1/N18-1101} {A broad-coverage
  challenge corpus for sentence understanding through inference}.
\newblock In \emph{Proceedings of the 2018 Conference of the North {A}merican
  Chapter of the Association for Computational Linguistics: Human Language
  Technologies, Volume 1 (Long Papers)}, pages 1112--1122, New Orleans,
  Louisiana. Association for Computational Linguistics.

\bibitem[{Wiseman et~al.(2017)Wiseman, Shieber, and
  Rush}]{wiseman-etal-2017-challenges}
Sam Wiseman, Stuart Shieber, and Alexander Rush. 2017.
\newblock \href {https://doi.org/10.18653/v1/D17-1239} {Challenges in
  data-to-document generation}.
\newblock In \emph{Proceedings of the 2017 Conference on Empirical Methods in
  Natural Language Processing}, pages 2253--2263, Copenhagen, Denmark.
  Association for Computational Linguistics.

\bibitem[{Yoshikawa et~al.(2017)Yoshikawa, Noji, and
  Matsumoto}]{yoshikawa-etal-2017-ccg}
Masashi Yoshikawa, Hiroshi Noji, and Yuji Matsumoto. 2017.
\newblock \href {https://doi.org/10.18653/v1/P17-1026} {{A}* {CCG} parsing with
  a supertag and dependency factored model}.
\newblock In \emph{Proceedings of the 55th Annual Meeting of the Association
  for Computational Linguistics (Volume 1: Long Papers)}, pages 277--287,
  Vancouver, Canada. Association for Computational Linguistics.

\end{thebibliography}
\bibliographystyle{acl_natbib}

\clearpage
\appendix

\begin{figure*}[t]
    \begin{minipage}[t]{\hsize}
        \scriptsize\centering
        \input{figures/treetrans_ex1a.tex}
        \subcaption{}

        \input{figures/treetrans_ex1b.tex}
        \subcaption{}
    \end{minipage} 
    \caption{An example tree transformation process for \textit{exactly} $n$ \textit{times}, where $n$ is a number.
    (a) is transformed into (b).}
    \label{figure:example-tt}
\end{figure*}

\begin{table}[t]
    \small\centering
    \begin{tabular}{lc} \toprule
Knowledge injection & Accuracy \\ \midrule
disabled & 0.23 \\
enabled  & \textbf{0.34} \\\bottomrule
\end{tabular}

    \caption{The accuracy of our proposed system with and without knowledge injection.}
    \label{table:result-ablation}
\end{table}

\section{Examples of Tree Transformation}
\label{section:example-tt}
We detect where to transform by tregex \citep{levy-andrew-2006-tregex}, the regular expression for trees.
We have three tsurgeon scripts, all of which are for handling numerical expressions involving the number of events.
For example, as Figure \ref{figure:example-tt} shows, we transform the CCG subtree (a) for \textit{exactly} $n$ \textit{times}, where $n$ is a number, into the CCG subtree (b).

\section{Ablation Study for Knowledge Injection}
\label{section:ablation-ki}
We conducted an ablation study for knowledge injection (see Section \ref{section:knowledge-injection}).
We picked all of the base hypotheses in our dataset (105 cases in total) and experimented to see how effective our knowledge injection method is.
As seen in Table \ref{table:result-ablation}, our knowledge injection method provided increased accuracy by 11\% (12 cases).

\end{document}